\documentclass{article}
\usepackage{ijcai19}

\usepackage{amssymb} 
\usepackage{tabularx}
\usepackage{balance}
\usepackage{natbib}

\usepackage{times}
\usepackage{soul}
\usepackage{url}
\usepackage[hidelinks]{hyperref}
\usepackage[utf8]{inputenc}
\usepackage[small]{caption}
\usepackage{graphicx}
\usepackage{amsmath}
\usepackage{booktabs}
\usepackage{xspace}
\usepackage{xcolor} 
\definecolor{ForestGreen}{rgb}{0.13, 0.55, 0.13}
\newcommand{\eegmamba}{ECG-RAMBA\xspace}

\title{\eegmamba: Zero-Shot ECG Generalization by
Morphology–Rhythm Disentanglement and
Long-Range Modeling}

\author{
Hai Duong Nguyen$^1$\and
Xuan-The Tran$^2$\footnote{\textit{Corresponding Author}}\\
\affiliations
$^1$HAI-Smartlink Research Lab, Anchi STE Company, Haiphong, Vietnam\\
$^2$School of Mechanical Engineering, Vietnam Maritime University, Haiphong, Vietnam
\footnote{\textit{Preprint notice:} This work has been submitted to the IEEE for possible publication. Copyright may be transferred without notice, after which this version may no longer be accessible.}
}

\begin{document}

\maketitle

\begin{abstract}
Deep learning has achieved strong performance for electrocardiogram (ECG) classification within individual datasets, yet dependable generalization across heterogeneous acquisition settings remains a major obstacle to clinical deployment and longitudinal monitoring. A key limitation of many model architectures is the implicit entanglement of morphological waveform patterns and rhythm dynamics, which can promote shortcut learning and amplify sensitivity to distribution shifts. We propose ECG-RAMBA, a framework that separates morphology and rhythm and then re-integrates them through context-aware fusion. ECG-RAMBA combines: (i) deterministic morphological features extracted by MiniRocket, (ii) global rhythm descriptors computed from heart-rate variability (HRV), and (iii) long-range contextual modeling via a bi-directional Mamba backbone. To improve sensitivity to transient abnormalities under windowed inference, we introduce a numerically stable Power Mean pooling operator ($Q=3$) that emphasizes high-evidence segments while avoiding the brittleness of max pooling and the dilution of averaging. We evaluate under a protocol-faithful setting with subject-level cross-validation, a fixed decision threshold, and no test-time adaptation. On the Chapman--Shaoxing dataset, ECG-RAMBA achieves a macro ROC-AUC $\approx 0.85$. In zero-shot transfer, it attains PR-AUC $=0.708$ for atrial fibrillation detection on the external CPSC-2021 dataset, substantially outperforming a comparable raw-signal Mamba baseline, and shows consistent cross-dataset performance on PTB-XL. Ablation studies indicate that deterministic morphology provides a strong foundation, while explicit rhythm modeling and long-range context are critical drivers of cross-domain robustness.
\end{abstract}

\section{Introduction}
Cardiovascular diseases (CVDs) remain the leading cause of death globally, underscoring the need for reliable automated tools that support early detection and longitudinal monitoring at scale~\cite{who_cvd_2025}. The electrocardiogram (ECG) is a cornerstone diagnostic test for assessing cardiac electrical activity and is routinely used to evaluate rhythm and conduction abnormalities in both acute and ambulatory settings~\cite{kligfield2007standardization}.

Recent advances in deep learning have pushed automated ECG interpretation to high performance on curated benchmarks and large-scale clinical datasets~\cite{hannun2019cardiologist,ribeiro2020automatic,zheng202012,cpsc2021}. Despite these advances, a persistent gap remains between strong in-distribution test results and reliable deployment in heterogeneous real-world environments. In practice, ECG data can differ substantially across institutions and time due to differences in acquisition hardware, lead placement, sampling rates, filtering, noise conditions, and patient populations~\cite{kligfield2007standardization}. Models that appear highly accurate on a single dataset may therefore degrade under distribution shift, limiting clinical trust and inhibiting longitudinal use.

A key reason for this brittleness is that many model architectures implicitly entangle multiple physiological factors within a single representation. Clinically, ECG interpretation integrates at least two complementary information streams: \textit{morphology} (e.g., P/QRS/T shape, ST--T deviations) and \textit{rhythm} (e.g., beat-to-beat variability and regularity). Rhythm is often summarized through heart rate variability (HRV) descriptors derived from R--R intervals, with standardized definitions and measurement practices established in the clinical literature~\cite{taskforce1996hrv}. In contrast, many end-to-end deep networks learn morphology and rhythm jointly from raw ECG signals \cite{hannun2019cardiologist,ribeiro2020automatic}, which can encourage shortcut learning based on dataset-specific acquisition cues and obscure which physiological factors drive predictions \cite{geirhos2020shortcut,ly2024shortcut,huang2024generalization}.

Motivated by this clinical separation, we introduce \textbf{ECG-RAMBA}, a physiologically informed framework that explicitly decouples morphological and rhythm cues and then re-integrates them through context-aware modeling. ECG-RAMBA uses \textbf{MiniRocket} to extract stable, deterministic morphological representations efficiently~\cite{dempster2021minirocket}, augments them with explicit \textbf{HRV} statistics to encode global rhythm dynamics~\cite{taskforce1996hrv}, and applies a \textbf{bi-directional Mamba} \cite{gu2024mamba} backbone to refine long-range temporal context with linear-time sequence modeling. This design aims to improve robustness under acquisition heterogeneity by (i) reducing over-reliance on raw-signal idiosyncrasies, and (ii) enforcing an interpretable division of labor between waveform morphology, rhythm descriptors, and long-range contextual evidence.

Finally, we note that evaluation choices in the ECG literature can substantially influence reported results. Protocols that allow subject overlap between training and test splits, or that tune decision thresholds on validation data in a way that does not reflect deployment constraints, can yield optimistic estimates. To better reflect real-world use, we adopt a strictly \textit{protocol-faithful} evaluation strategy with subject-level separation and fixed decision rules across experiments. Under this setting, we consistently observe a \textbf{ranking--decision gap}—strong ranking metrics (e.g., ROC-AUC) paired with more conservative fixed-threshold operating performance—which we interpret as an expected and safety-aligned behavior for clinical decision support systems.

In summary, our main contributions are:
\begin{itemize}
    \item \textbf{Physiologically grounded architecture:} We propose a multi-view framework that integrates deterministic morphological extraction (MiniRocket), explicit rhythm modeling via HRV, and efficient long-range contextual refinement with a Mamba backbone.
    \item \textbf{Robust zero-shot generalization:} We show that ECG-RAMBA generalizes to unseen datasets without fine-tuning, outperforming a comparable raw-signal Mamba-based baseline for atrial fibrillation detection on external data and demonstrating consistent cross-dataset behavior.
    \item \textbf{Protocol-faithful evaluation:} We establish a deployment-oriented evaluation protocol with subject-aware splits, fixed decision thresholds, and no test-time adaptation, and analyze the resulting ranking--decision gap as a clinically appropriate conservatism signal rather than a modeling deficiency.
\end{itemize}

\section{Related Work}
\label{sec:related_work}

\subsection{Deep Learning for ECG Analysis}
Deep learning has substantially advanced automated ECG interpretation, with large-scale studies demonstrating near-expert performance for arrhythmia and abnormality detection under controlled evaluation settings~\cite{hannun2019cardiologist,ribeiro2020automatic}. This progress has been accelerated by the availability of public 12-lead ECG resources and benchmarks, including the Chapman--Shaoxing database~\cite{zheng202012} and PTB-XL~\cite{wagner2020ptbxl}, as well as community challenges that aggregate multi-institutional cohorts to encourage reproducible evaluation~\cite{cpsc2021}.

Despite strong in-distribution results, robust deployment remains limited by \emph{dataset shift}: ECG recordings can differ markedly across institutions due to device characteristics, sampling rates, filtering pipelines, lead configurations, annotation conventions, and patient demographics. Recent reviews and analyses emphasize that performance can drop substantially when models are transferred to external cohorts without adaptation, highlighting a persistent generalization gap in clinical ECG AI~\cite{huang2024generalization,hong2022practical}. More broadly, the medical AI community has documented that end-to-end models can exploit \emph{shortcuts}---spurious or dataset-specific cues that correlate with labels in the training distribution but do not reflect causal physiology---leading to brittle behavior under distribution shifts~\cite{ly2024shortcut, nguyen2024fairad}.

Architecturally, CNNs remain a strong baseline for ECG modeling due to their inductive bias for local waveform structure, while attention-based models are increasingly explored for capturing longer temporal context, especially in multi-label 12-lead settings~\cite{natarajan2020wide,vaswani2017attention}. However, quadratic attention complexity can impose practical constraints on long recordings, often requiring truncation, aggressive downsampling, or windowed inference, which may reduce sensitivity to subtle yet clinically relevant morphology~\cite{vaswani2017attention,hong2022practical}. These observations motivate approaches that explicitly target cross-domain robustness rather than solely optimizing within-dataset metrics.

\subsection{State Space Models for Efficient Long-Range Temporal Modeling}

Structured State Space Models (SSMs) have emerged as efficient alternatives to attention for long-sequence learning. S4-style models formalize sequence processing via structured continuous-time state dynamics, enabling long-range memory with linear-time inference in sequence length~\cite{gu2022s4}. More recently, Mamba introduces selective state-space mechanisms that further improve practical performance while retaining linear scaling, making SSMs attractive backbones for long physiological time series~\cite{gu2024mamba, tran2024eeg}.

For ECG, long-range temporal structure is clinically important for rhythm disorders whose evidence may unfold over many beats (e.g., irregularity patterns and episode-level dynamics). Nonetheless, purely sequence-centric backbones may not, by themselves, impose strong inductive structure for fine-grained morphological cues (e.g., QRS notching, ST--T deviations) that are typically well captured by localized filters and shape-sensitive representations. This motivates hybrid designs that combine morphology-specialized encoders with efficient long-context refinement.

\subsection{Hybrid and Physiologically Informed Designs}

Clinical ECG interpretation relies on complementary information streams: waveform \emph{morphology} (shape and segment-level deviations) and \emph{rhythm} (beat-to-beat timing and variability). Rhythm descriptors derived from heart rate variability (HRV) have long been standardized and widely used as summary markers of autonomic and rhythm dynamics~\cite{taskforce1996hrv}. In parallel, time-series classification research has developed strong deterministic feature transforms that capture diverse local patterns with minimal training overhead. ROCKET and MiniRocket use random convolutional kernels and simple pooling statistics to produce highly effective, numerically stable morphological representations for time series~\cite{dempster2020rocket,dempster2021minirocket}.

However, deterministic transforms alone typically lack mechanisms to model evolving temporal context across windows or across beats, while monolithic end-to-end models that learn morphology and rhythm jointly from raw signals can be more susceptible to shortcut learning and domain-specific confounds~\cite{ly2024shortcut,hong2022practical}. 

\textbf{ECG-RAMBA} integrates these complementary directions by (i) extracting stable morphology features using MiniRocket, (ii) encoding explicit rhythm information via HRV statistics grounded in clinical standards, and (iii) refining long-range temporal context with a bi-directional SSM backbone inspired by recent advances in efficient sequence modeling. This explicit separation-and-fusion design is motivated by the hypothesis that \emph{physiological disentanglement} can reduce reliance on dataset-specific cues and improve zero-shot robustness across heterogeneous acquisition environments.

\begin{figure*}[t]
    \centering
    \includegraphics[width=\textwidth]{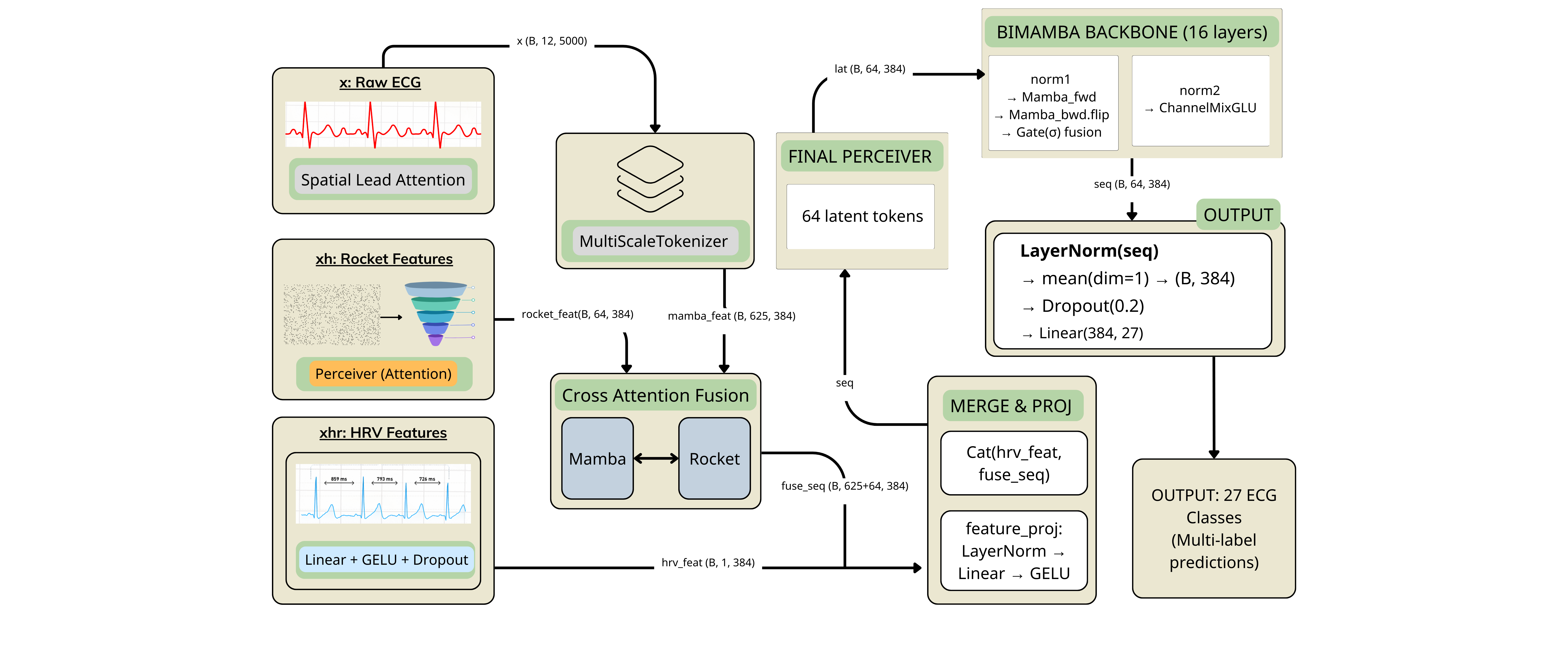}
    \caption{Overview of the proposed ECG-RAMBA architecture, illustrating the integration of deterministic morphological features (MiniRocket), rhythm-related descriptors (HRV), and bidirectional state-space modeling (BiMamba) within a unified, protocol-faithful framework.}
    \label{fig:architecture}
\end{figure*}

\section{Methodology}
\label{sec:methodology}

The proposed framework, ECG-RAMBA, is designed to emulate the clinical ECG interpretation workflow by explicitly decoupling morphological analysis from rhythm interpretation before synthesizing them in a context-aware manner. This design is motivated by the clinical observation that waveform morphology and temporal rhythm constitute complementary but physiologically distinct diagnostic cues. The overall architecture is illustrated in Fig.~\ref{fig:architecture}.

\subsection{Signal Preprocessing and Event-Centric Slicing}

Raw ECG signals $X \in \mathbb{R}^{12 \times N}$ (sampled at 500 Hz) are processed to suppress noise while preserving physiological fidelity. We apply a 4th-order Butterworth bandpass filter with cutoffs at $[0.5, 40]$ Hz to remove baseline wander and myoelectric artifacts.

Crucially, to prevent \textit{distributional leakage} during normalization, we employ \textbf{Instance-wise Z-score Normalization}. For each lead $c$ of a recording $x$, normalization is performed using only the statistics of that specific instance:
\begin{equation}
    x_{c,norm} = \frac{x_c - \mu(x_c)}{\sigma(x_c) + \epsilon}
\end{equation}
This design explicitly prevents global dataset statistics from leaking into individual samples, thereby mitigating reliance on dataset-level statistical artifacts and ensuring subject-independent evaluation.

To handle variable-length recordings and capture transient arrhythmias (e.g., paroxysmal AF), we adopt an event-centric slicing strategy. Recordings are decomposed into overlapping windows of length $L=2500$ (5 seconds) with a stride $S=1250$ (50\% overlap). This yields a bag of slices $\mathcal{V}_i = \{x_{i,j}\}_{j=1}^{M_i}$, where each slice inherits the parent record's label.

\subsection{Physiologically Informed Feature Engineering}

We inject explicit clinical inductive biases through two specialized feature streams to disentangle morphology from rhythm, reflecting the dual-pathway reasoning routinely employed by cardiologists during ECG interpretation.

\subsubsection{Deterministic Morphological Features (MiniRocket)}

Deep CNNs trained end-to-end on limited ECG datasets are prone to overfitting high-frequency noise and dataset-specific artifacts. To mitigate this, we utilize MiniRocket~\cite{dempster2021minirocket}, a shallow random convolutional network that extracts 10,000 deterministic features per slice without gradient-based optimization. These features are fixed throughout training, providing a stable morphological representation.

To handle the high dimensionality, we apply a \textbf{Fold-Aware Principal Component Analysis (PCA)}. Specifically, the projection matrix $W_{\text{PCA}}$ is computed independently for each training fold and applied to validation/test folds, ensuring strict separation between training and evaluation data. The features are projected to $D_{\text{morph}}=3072$, retaining more than $99\%$ of the explained variance on the training data while reducing the feature space by approximately $70\%$. This design balances morphological expressiveness with computational efficiency while preventing information leakage.

\subsubsection{Global Rhythm Statistics (HRV)}

Since event-centric slicing disrupts long-range heart rate variability patterns, we extract a global rhythm vector $h_{\text{hrv}} \in \mathbb{R}^{36}$ from the full-length recording. This vector includes time-domain metrics (e.g., RMSSD, SDNN), frequency-domain indices (e.g., LF/HF ratio), and higher-order statistical moments (e.g., kurtosis, skewness).

The rhythm vector is broadcast as a fixed conditioning vector to all slices within the same recording, providing a static global context $C_{\text{global}}$ that complements local morphological evidence. This design preserves rhythm information that cannot be reliably recovered from short temporal windows alone, particularly for irregular or paroxysmal arrhythmias.

\subsection{ECG-RAMBA Architecture}

The extracted features are integrated through a learnable backbone designed for efficient long-range temporal modeling and physiologically grounded feature fusion.

\subsubsection{Spatial Lead Attention}
\label{subsubsec:spatial_attn}

To explicitly model inter-lead dependencies and improve robustness to noisy or uninformative channels, we introduce a \textit{Spatial Lead Attention} module prior to tokenization. Each lead $l \in [1,12]$ is summarized into a temporal embedding $e_l$ via pooling, combined with a learnable lead-identity encoding, and processed using self-attention to produce a scalar gating weight $\alpha_l \in [0,1]$.

The re-weighted signal $\tilde{x}_l = \alpha_l \cdot x_l$ allows the model to dynamically prioritize diagnostically relevant leads (e.g., precordial leads for conduction abnormalities such as bundle branch blocks) while suppressing noisy or redundant channels, aligning with established lead-specific clinical reasoning.

\subsubsection{Multi-Scale Tokenizer}

The spatially re-weighted signal is tokenized using a multi-branch convolutional tokenizer to capture patterns at multiple temporal resolutions. We employ three parallel convolutional branches with kernel sizes $k \in \{3,7,15\}$, corresponding to fine-grained (QRS complexes) and coarse-grained (T-wave) structures. The outputs are concatenated and projected to form a token sequence $T \in \mathbb{R}^{L' \times D_{\text{model}}}$, where $D_{\text{model}}=384$.

\subsubsection{Bi-Directional Mamba Backbone}

We employ a Bi-Directional State Space Model (Mamba) to capture long-range temporal dependencies with linear complexity. The backbone consists of $N=16$ stacked Mamba blocks with a model dimension $D_{\text{model}}=384$ and an inner-state expansion factor of 2.

The continuous-time dynamics $h'(t) = \mathbf{A}h(t) + \mathbf{B}x(t)$ are discretized using Zero-Order Hold (ZOH) with time-scale parameter $\Delta$:
\begin{align}
    \overline{\mathbf{A}} &= \exp(\Delta \mathbf{A}), \quad
    \overline{\mathbf{B}} = (\Delta \mathbf{A})^{-1}(\exp(\Delta \mathbf{A}) - \mathbf{I}) \cdot \Delta \mathbf{B}, \\
    h_t &= \overline{\mathbf{A}} h_{t-1} + \overline{\mathbf{B}} x_t.
\end{align}

While the present study focuses on fixed-length clinical recordings, the choice of a state space backbone is motivated by its ability to scale linearly to long and continuous ECG streams, such as Holter or wearable monitoring, where attention-based architectures become memory-prohibitive.

To incorporate both past and future context—critical for ECG interpretation—we adopt a bi-directional formulation:
\begin{equation}
    H_{\text{fwd}} = \text{Mamba}(T), \quad
    H_{\text{bwd}} = \text{flip}(\text{Mamba}(\text{flip}(T))).
\end{equation}
The two directions are fused via a learnable gating mechanism to produce the contextual representation $H_{\text{ctx}}$.

\subsubsection{Cross-Modal Fusion with Attention}

To synthesize heterogeneous features, we employ Multi-Head Cross-Attention with $H=8$ heads. The contextual sequence $H_{\text{ctx}}$ acts as the Query ($Q$), while the concatenated static features $(h_{\text{morph}} \oplus h_{\text{hrv}})$ serve as the Key ($K$) and Value ($V$):
\begin{equation}
    \text{Attention}(Q,K,V) = \text{softmax}\left(\frac{QK^{T}}{\sqrt{d_k}}\right)V.
\end{equation}

Unlike static gating or heuristic fusion schemes, cross-attention enables \textit{time-dependent physiological routing}, allowing the model to emphasize morphological cues during waveform-critical segments and rhythm statistics during temporally irregular intervals. This design reflects the non-stationary nature of ECG interpretation and avoids imposing a fixed global importance between modalities.

\subsection{Numerically Stable Slice-to-Record Aggregation}
\label{subsec:aggregation}

To aggregate slice-level probabilities $\mathcal{P} = \{\hat{p}_{i,j}\}_{j=1}^{M_i}$ into a record-level prediction $\hat{P}_i$, we employ \textbf{Power Mean Pooling} with exponent $Q=3$:
\begin{equation}
    \hat{P}_i = \left(\frac{1}{M_i} \sum_{j=1}^{M_i} (\hat{p}_{i,j})^Q\right)^{1/Q}.
\end{equation}

From a physiological perspective, many clinically relevant arrhythmias are transient events that manifest in only a small fraction of a long recording. For example, in paroxysmal atrial fibrillation, brief irregular episodes may be sufficient for diagnosis. Power Mean Pooling functions as a \textit{non-linear confident aggregation} operator: high-confidence pathological slices contribute super-linearly, while low-confidence background segments are suppressed. Setting $Q=3$ provides a principled trade-off between sensitivity to rare events and robustness to noise, avoiding both the brittleness of max pooling and the dilution effect of average pooling.

To ensure numerical robustness under mixed-precision inference and long recordings, aggregation is computed in the log-domain using the LogSumExp trick:
\begin{equation}
    \log(\hat{P}_i) = \frac{1}{Q}\left[\log \sum_{j=1}^{M_i} \exp\left(Q \cdot \log(\hat{p}_{i,j})\right) - \log(M_i)\right].
\end{equation}
Probabilities are clamped to $[\epsilon, 1-\epsilon]$ ($\epsilon=10^{-6}$) prior to transformation, ensuring stable and production-ready behavior.

\subsection{Training Protocol}

We enforce a strictly protocol-faithful training regime:
\begin{itemize}
    \item \textbf{Loss Function:} Training is warmed up with Binary Cross-Entropy (BCE) for 8 epochs, followed by Asymmetric Loss (ASL) with $\gamma_- = 2.5, \gamma_+ = 0$ to address class imbalance without oversampling.
    \item \textbf{Optimization:} AdamW optimizer with peak learning rate $9 \times 10^{-4}$ and cosine annealing to $10^{-6}$ is used. A batch size of 192 slices is employed, while maintaining strict subject-level separation.
    \item \textbf{Robustness:} Final evaluation uses Exponential Moving Average (EMA) of model weights with decay $0.999$.
\end{itemize}

\subsection{Theoretical and Practical Efficiency}

A key advantage of the Mamba backbone over Transformer-based architectures lies in its linear scaling with respect to sequence length $L$. For high-resolution ECG signals ($L = 5000$), standard self-attention mechanisms incur a quadratic complexity of $\mathcal{O}(L^2)$, leading to substantial memory overhead and limiting their applicability to long physiological recordings. In contrast, ECG-RAMBA leverages the selective scan mechanism of state space models, achieving linear complexity $\mathcal{O}(L)$ and enabling efficient long-range temporal modeling without aggressive downsampling.

Beyond theoretical complexity, practical efficiency in ECG-RAMBA is achieved through architectural modularization. Fine-grained morphological information is extracted using the frozen MiniRocket module, while learnable components are reserved for contextual integration and decision making. This separation reduces the number of trainable parameters compared to fully learnable deep CNN ensembles, improving optimization stability and mitigating overfitting on limited clinical datasets, while preserving sensitivity to subtle electrophysiological patterns.

\section{Experiments and Results}
\label{sec:experiments}

\subsection{Datasets and Experimental Setup}
We evaluate ECG-RAMBA across three publicly available ECG datasets to assess both in-distribution performance and real-world generalization.

\textbf{Chapman–Shaoxing (CS)} serves as the primary dataset for training and in-distribution evaluation, comprising over 45,000 12-lead ECG recordings sampled at 500\,Hz. To eliminate subject leakage—a pervasive issue in prior benchmarks—we strictly enforce a \textbf{5-fold subject-aware cross-validation} protocol. Diagnostic classes with extremely low support are retained to reflect realistic long-tail label distributions.

To assess cross-dataset robustness, we conduct zero-shot evaluation on two external cohorts: \textbf{CPSC-2021}, which challenges the model with paroxysmal atrial fibrillation under significant device noise and sampling shifts; and \textbf{PTB-XL}, which introduces diverse diagnostic superclasses with distinct acquisition protocols and lead configurations.

Crucially, we adhere to a \textbf{protocol-faithful evaluation} framework, 
in which a fixed decision threshold ($\tau = 0.5$) is applied uniformly across all experiments. 
No validation-time threshold selection, post-hoc decision adjustment, 
or test-time adaptation is employed. 
This evaluation design is intended to reflect deployment-oriented conditions, 
prioritizing realistic clinical behavior over optimistically tuned benchmark performance.

\subsection{In-Distribution Performance}

\subsubsection{Overall Performance and the ``Safety Gap''}
Table~\ref{tab:in_distribution} summarizes the in-distribution results on the Chapman–Shaoxing dataset. Regarding the aggregation hyperparameter, sensitivity analysis indicated that while higher exponents ($Q=8$) marginally improve F1, they increase sensitivity to noise artifacts. We therefore adopted \textbf{$Q=3$} as a robust operating point, functioning as a ``soft-attention'' mechanism that emphasizes pathological events without compromising numerical stability.

ECG-RAMBA achieves a mean \textbf{Macro ROC-AUC of $0.848 \pm 0.047$} and a \textbf{Macro F1-score of $0.312 \pm 0.052$}. The significant divergence between the high discrimination capability (ROC-AUC) and the moderate fixed-threshold classification metric (F1) is a critical finding. We interpret this ``Ranking--Decision Gap'' not as a performance deficit, but as a desirable \textbf{safety characteristic} for clinical decision support systems. The model maintains a \textit{conservative prediction bias}, refusing to issue high-confidence alerts for ambiguous cases to minimize false positive and mitigate ``alarm fatigue'' for clinicians.

\begin{table}[h]
\caption{In-distribution performance on Chapman–Shaoxing using 5-fold subject-aware cross-validation. Metrics are reported at a fixed threshold $\tau=0.5$. The ``Ranking–Decision Gap'' reflects a safety-oriented bias in predictions.}
\label{tab:in_distribution}
\centering
\begin{tabular}{lcc}
\hline
\textbf{Metric} & \textbf{Mean} & \textbf{Std} \\
\hline
Macro F1 & 0.3115 & 0.0524 \\
Macro ROC-AUC & 0.8479 & 0.0471 \\
Macro PR-AUC & 0.2914 & 0.0689 \\
\hline
\end{tabular}
\end{table}

\subsubsection{Per-Class Analysis}
\begin{figure*}[t]
    \centering
    \includegraphics[width=\linewidth]{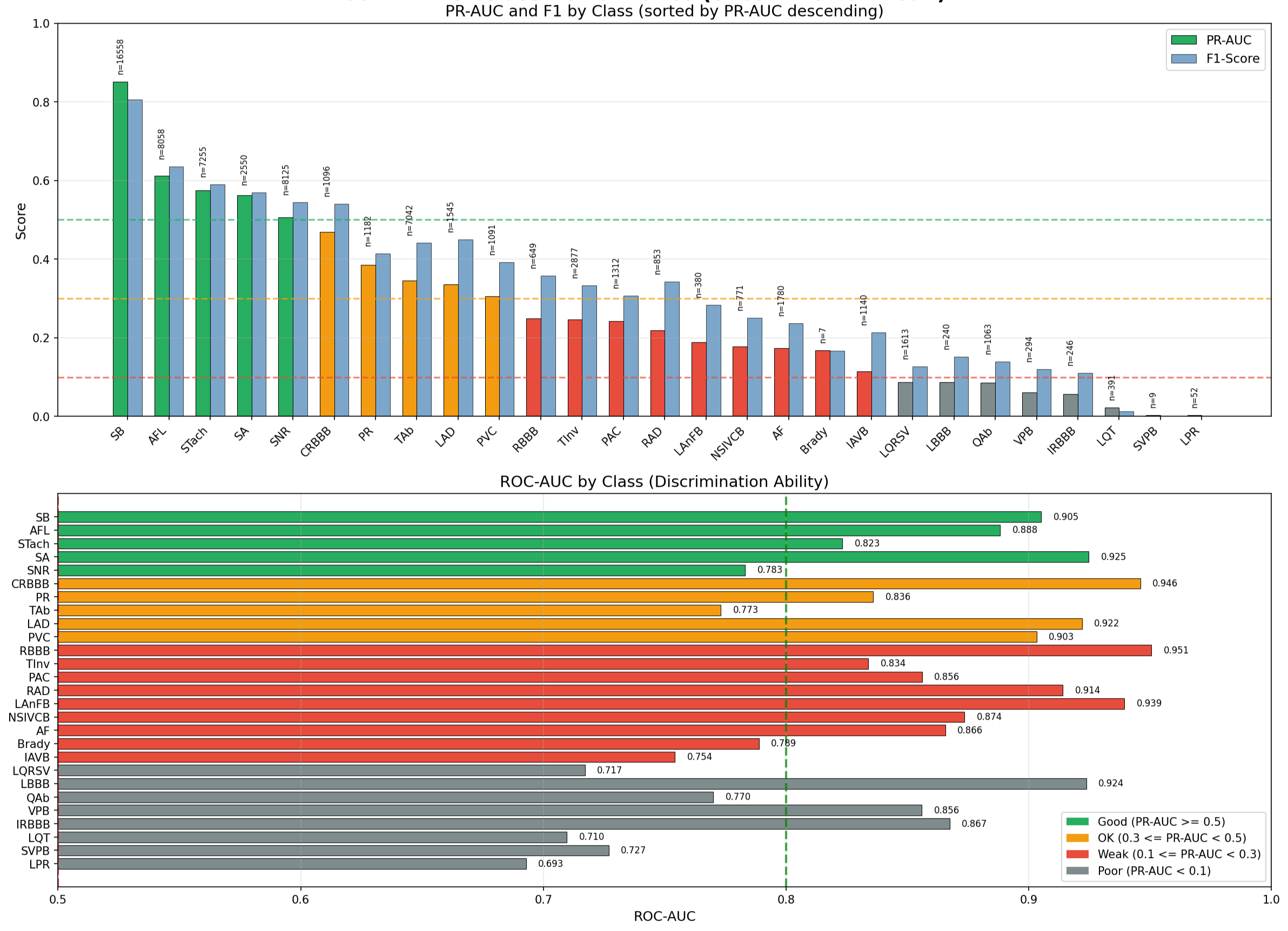} 
    \caption{Per-class performance comparison. The discrepancy between high ROC-AUC (orange) and conservative F1-scores (blue) highlights the model's safety-oriented ranking capability, particularly for morphological abnormalities.}
    \label{fig:per_class}
\end{figure*}

Fig.~\ref{fig:per_class} reveals a distinct performance dichotomy grounded in physiology. \textbf{Rhythm-dominant classes} (e.g., Sinus Bradycardia, Atrial Flutter) achieve consistently strong F1-scores, validating the model's ability to capture temporal periodicities. Conversely, \textbf{Morphology-dominant classes} (e.g., Bundle Branch Blocks, ST-segment changes) exhibit high ROC-AUC but lower F1-scores, indicating reliable ranking capability even when absolute probability assignment remains conservative.



\subsection{Protocol-Faithful Evaluation and Behavioral Validation}

Direct numerical comparison with previously reported ECG benchmarks is inherently misleading due to substantial differences in evaluation protocols. Many high-performing results in the literature rely on random or record-level splits, which may inadvertently introduce subject-level information leakage and inflate performance. In contrast, ECG-RAMBA is evaluated under a strictly subject-aware cross-validation protocol with fixed decision thresholds, providing a deployment-faithful estimate of achievable performance.

Rather than emphasizing leaderboard-style comparisons, we focus on validating whether the proposed architecture exhibits clinically and physiologically consistent behavior under controlled perturbations and real-world distribution shifts. Two complementary analyses support this validation.

First, zero-shot evaluation on external datasets reveals a clear dichotomy between rhythm-dominant and morphology-dominant disorders. Rhythm abnormalities, such as paroxysmal atrial fibrillation on CPSC-2021, generalize robustly across acquisition devices and noise conditions, confirming that the model captures global temporal invariants. In contrast, morphology-dependent pathologies, particularly myocardial infarction on PTB-XL, experience performance degradation under cross-dataset transfer, reflecting sensitivity to lead-set configuration.

Second, the structured lead dropout study provides controlled evidence of physiological disentanglement. Removing precordial leads causes a substantial degradation in morphology-driven performance, while rhythm performance remains stable or slightly improves. This behavior aligns with established clinical knowledge and confirms that the model relies on appropriate physiological cues rather than spurious dataset correlations.

Together, these analyses demonstrate that ECG-RAMBA prioritizes robust, interpretable behavior under realistic evaluation conditions, offering insights that extend beyond conventional benchmark comparisons.

\subsection{Zero-Shot Generalization}

\subsubsection{Rhythm Robustness (CPSC-2021)}
We evaluate robustness to device heterogeneity using the CPSC-2021 dataset for Paroxysmal Atrial Fibrillation detection. As shown in Fig.~\ref{fig:cpsc_zeroshot}, our proposed hybrid framework (Branch A) substantially outperforms the raw-signal baseline (Branch B). ECG-RAMBA achieves a \textbf{PR-AUC of 0.708}, representing an approximately \textbf{85\%} relative improvement over the raw-signal Mamba baseline (0.382). This underscores a key insight: explicit, non-learnable rhythm features (HRV) serve as an invariant ``anchor,'' allowing the model to bridge the domain gap when learnable features are perturbed by sensor noise.

\begin{figure}[t]
    \centering
    \includegraphics[width=0.8\linewidth]{fig_cpsc_combined.pdf}
    \caption{Zero-shot generalization on CPSC-2021. The proposed method (Branch A) maintains robust AF detection, whereas the raw-signal baseline (Branch B) degrades significantly.}
    \label{fig:cpsc_zeroshot}
\end{figure}

\subsubsection{Spatial Dependency in Zero-Shot (PTB-XL)}
\begin{figure}[t]
    \centering
    \includegraphics[width=0.8\linewidth]{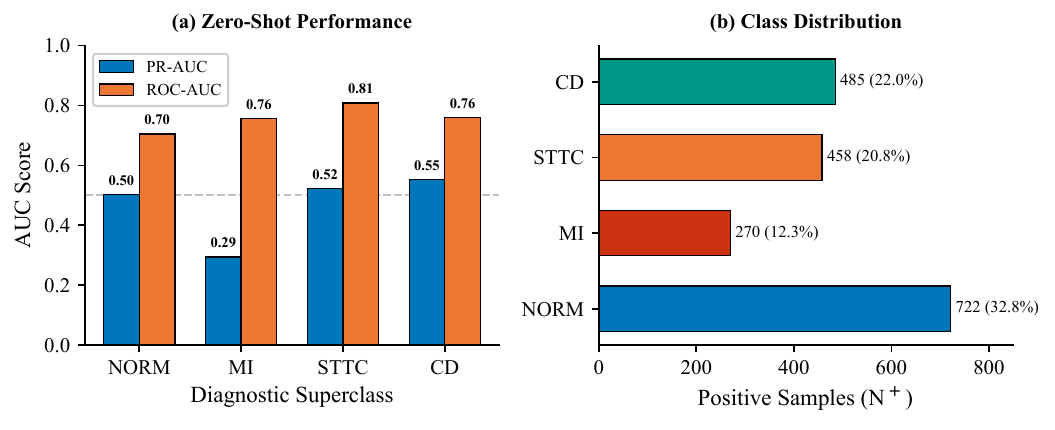}
    \caption{Zero-shot performance on PTB-XL. Global patterns (CD, STTC) transfer well, while localized pathologies (MI) are impacted by lead-set variations.}
    \label{fig:ptb_zeroshot}
\end{figure}

Fig.~\ref{fig:ptb_zeroshot} presents generalization on PTB-XL. The model effectively transfers to global patterns such as Conduction Disturbances (CD) and ST-T changes. However, performance degrades for Myocardial Infarction (MI). This is physiologically consistent: MI diagnosis relies on precise vectorcardiographic criteria (e.g., localized ST-elevation), which are highly sensitive to the inevitable variations in electrode placement between the Chapman and PTB-XL acquisition protocols.

\subsection{Spatial and Temporal Interpretability}
\label{sec:spatial_analysis}

\subsubsection{Physiological Disentanglement: A Lead Dropout Study}
\begin{figure}[h]
    \centering
    \includegraphics[width=\linewidth]{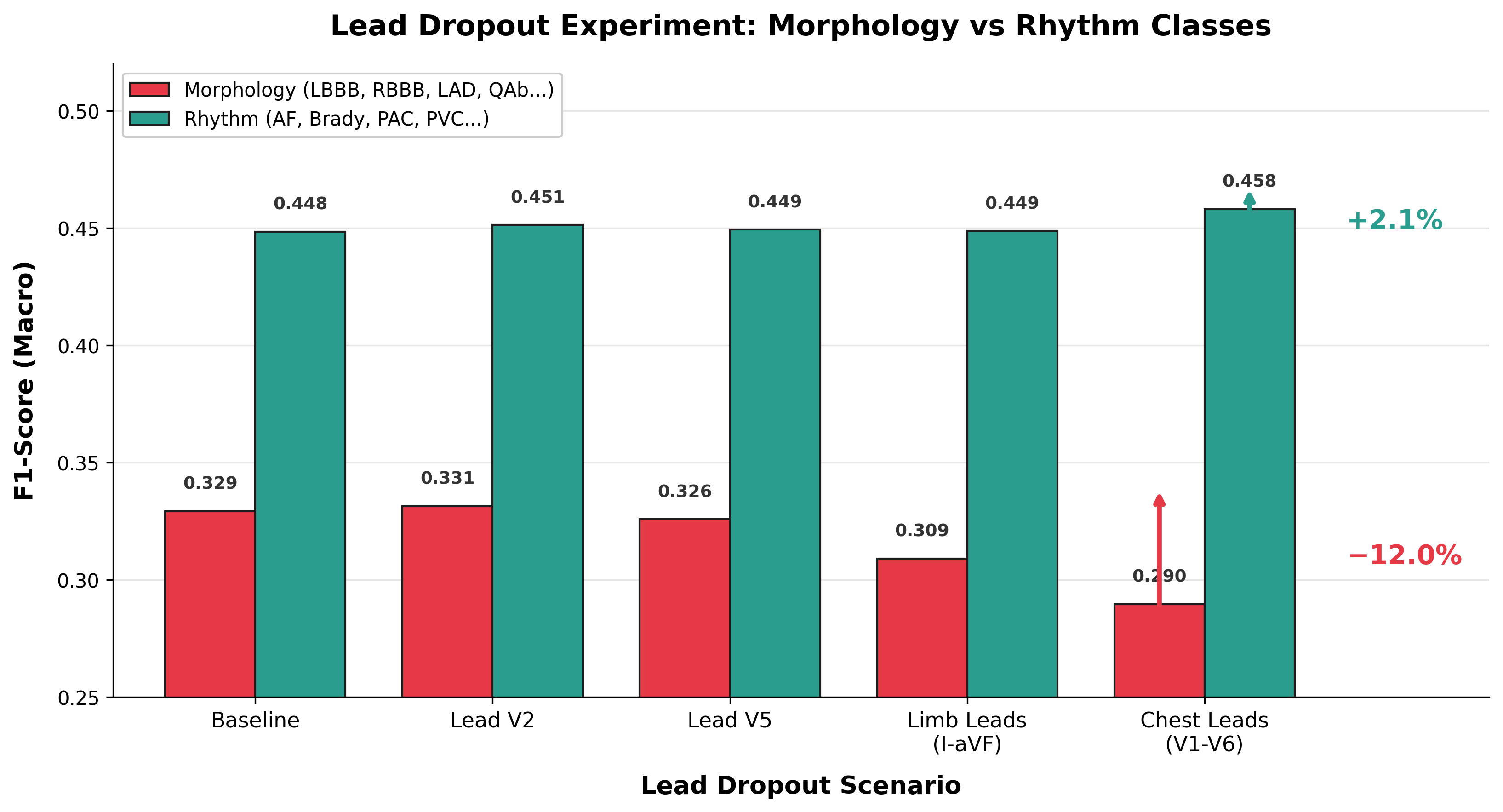}
    \caption{Impact of Lead Dropout on Morphology vs. Rhythm. Morphology classes (red) degrade significantly without spatial context (-12.0\%), while Rhythm classes (green) remain robust (+2.1\%), validating the physiological disentanglement hypothesis.}
    \label{fig:lead_dropout}
\end{figure}

To validate the model's internal logic, we performed a structural lead dropout experiment, masking precordial leads (V1--V6) and observing the impact on Morphology vs. Rhythm classes (Fig.~\ref{fig:lead_dropout}).

The results provide clear qualitative evidence of \textbf{physiological disentanglement}. When spatial context is removed:
\begin{itemize}
    \item \textbf{Morphology performance drops by 12.0\%}, confirming that the model correctly treats structural defects as spatially dependent vector quantities.
    \item \textbf{Rhythm performance increases by 2.1\%}. This counterintuitive gain suggests that the model treats rhythm as a temporal invariant. Removing spatial channels likely reduces ``morphological noise,'' forcing the attention mechanism to focus more sharply on the rhythm-encoding pathways (HRV/Mamba).
\end{itemize}

\subsubsection{Temporal Relevance (Saliency Analysis)}
\begin{figure}[h]
    \centering
    \includegraphics[width=\linewidth]{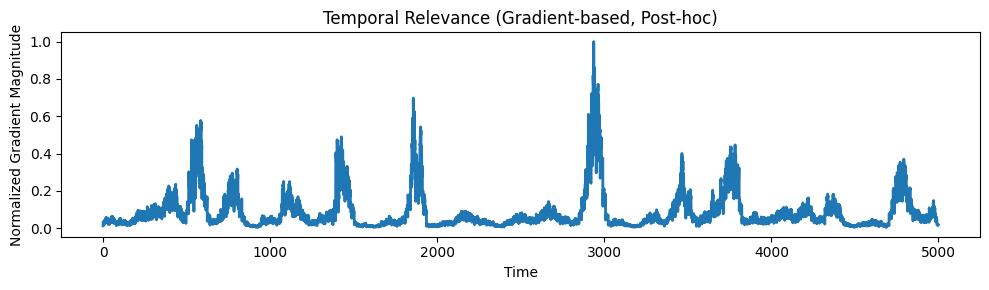} 
    \caption{Gradient-based Temporal Saliency Map. The Mamba backbone correctly attends to transient pathological events (high amplitude regions corresponding to arrhythmia) rather than background noise, demonstrating physiological grounding.}
    \label{fig:saliency}
\end{figure}

To assess whether the model relies on physiologically meaningful cues rather than spurious correlations, we visualized the gradient-based saliency map derived from the Mamba backbone (Fig.~\ref{fig:saliency}). The model selectively highlights clinically meaningful segments—specifically irregular R-R intervals in Atrial Fibrillation and abnormal QRS complexes—while suppressing baseline segments. This confirms that the event-centric slicing and Mamba's selective scan mechanism successfully capture long-range pathological dependencies.

\subsection{Computational Complexity and Efficiency}
Table~\ref{tab:complexity} summarizes the computational characteristics of ECG-RAMBA. A key architectural advantage is the use of the Bi-Directional Mamba backbone, which achieves linear complexity $\mathcal{O}(L)$ with respect to sequence length, overcoming the quadratic bottleneck $\mathcal{O}(L^2)$ of standard Transformers.

Although the total parameter count is relatively high (\textbf{72.83M}), this capacity is concentrated in the backbone to support deep contextual modeling of 5000-step sequences. Empirically, the model remains highly efficient with \textbf{10.77 GFLOPs} per forward pass. Real-world latency testing (batch size=1) yields an inference time of \textbf{65.05 ms} per full 12-lead record, confirming feasibility for real-time clinical deployment.

\begin{table}[h]
\caption{Computational Complexity of ECG-RAMBA. Metrics are reported for a single forward pass of a full 12-lead ECG record ($12 \times 5000$).}
\label{tab:complexity}
\centering
\renewcommand{\arraystretch}{1.25}
\begin{tabularx}{\columnwidth}{@{} l l X @{}}
\toprule
\textbf{Metric} & \textbf{Value} & \textbf{Note} \\
\midrule
Total Parameters & 72.83M & All trainable. BiMamba Backbone constitutes 86.9\% of capacity. \\
FLOPs (Forward) & 10.77 G & Lower bound estimate (actual compute may be 1.5--2$\times$ higher due to SSM operations). \\
Inference Latency & 65.05 ms & Batch size = 1, full record processing ($12 \times 5000$). \\
Training Memory & $\sim$12 GB & Measured on NVIDIA A100 GPU. \\
\bottomrule
\end{tabularx}
\end{table}

\subsection{Ablation Studies}

\begin{figure}[t]
    \centering
    \includegraphics[width=0.9\linewidth]{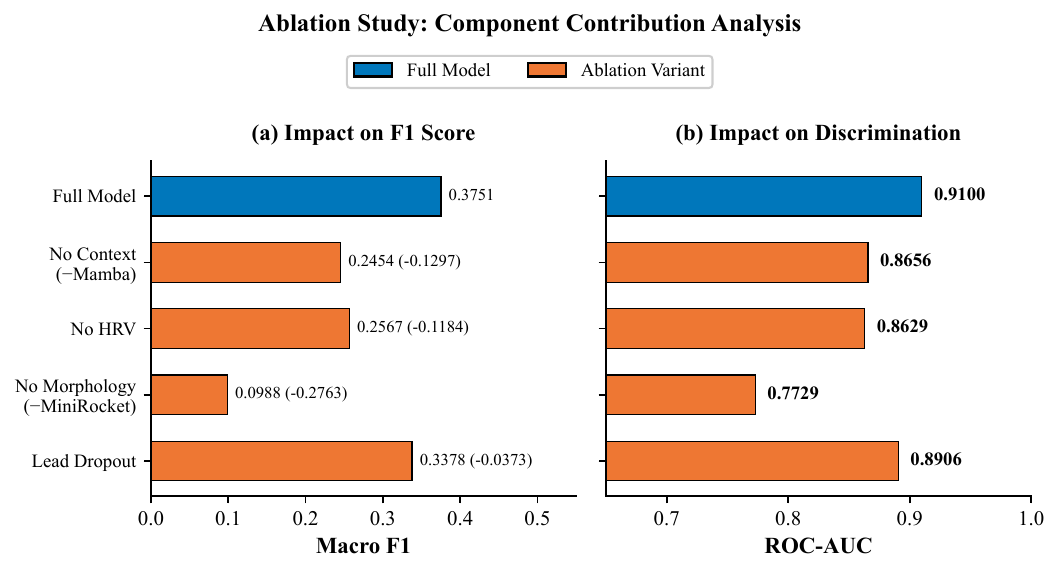} 
    \caption{Comprehensive ablation study. The significant performance drop when removing either the Context (Mamba) or Rhythm (HRV) component validates the necessity of the proposed multi-view design.}
    \label{fig:ablation}
\end{figure}

Fig.~\ref{fig:ablation} presents a controlled ablation analysis designed to isolate the functional contribution of individual architectural components. To ensure fair comparison across variants, all ablation results are evaluated under an identical validation setting rather than aggregated cross-validation; therefore, absolute values are not directly comparable to the cross-validated results reported in Table~I.

Removing the Mamba backbone leads to a pronounced degradation in Macro F1 (approximately $-0.13$), indicating that deterministic feature extraction alone is insufficient to integrate local cues into a coherent temporal representation. Similarly, excluding HRV features consistently reduces performance (approximately $-0.12$), reaffirming the importance of explicit rhythm statistics for capturing long-range temporal irregularities. The full ECG-RAMBA architecture achieves the strongest performance under this controlled setting, validating the complementary roles of deterministic morphological extraction and learnable contextual modeling. Importantly, performance degradation trends are consistent across all evaluated folds, indicating robust component-level contributions rather than fold-specific artifacts.

\section{Discussion}
\label{sec:discussion}

This work targets a persistent gap between benchmark-leading ECG classifiers and deployment-grade reliability by introducing ECG-RAMBA, a framework centered on (i) physiologically motivated disentanglement of rhythm and morphology and (ii) protocol-faithful evaluation intended to reflect realistic clinical use. Beyond aggregate metrics, our analyses highlight how distinct clinical cues behave under distribution shift, clarifying which failure modes are structural (i.e., inherent to ECG acquisition and clinical criteria) versus algorithmic.

\subsection{Physiological Disentanglement: Rhythm vs. Morphology}

Our zero-shot results reveal a clinically consistent dichotomy between rhythm-dominant and morphology-dominant conditions. For rhythm disorders, ECG-RAMBA exhibits strong cross-dataset transferability. In particular, on CPSC-2021, the proposed model substantially improves paroxysmal atrial fibrillation detection over a raw-signal baseline, supporting the view that combining explicit rhythm descriptors (HRV) with long-range sequence modeling yields a more portable representation under heterogeneous acquisition.

In contrast, zero-shot generalization for morphology-dominant conditions is more nuanced. While transfer remains relatively stable for broader ST--T abnormalities, performance degrades for spatially localized pathologies such as myocardial infarction (MI) on PTB-XL (Fig.~\ref{fig:ptb_zeroshot}). This behavior is physiologically expected: standard diagnostic criteria for ST-elevation MI rely on ST-segment changes in \emph{specific} (often contiguous) leads, reflecting regional injury patterns and lead-specific projections of the cardiac electrical vector \cite{thygesen2018udmi}. Consequently, differences in lead placement and acquisition protocols can act as structured spatial perturbations that alter measured morphology even when underlying physiology is unchanged \cite{finlay2015electrode,gregory2021chestplacement}.

This interpretation is supported by the lead dropout analysis (Fig.~\ref{fig:lead_dropout}). Morphology-dominant classes show pronounced sensitivity when precordial leads are removed, whereas rhythm-dominant performance remains comparatively stable. Together, these findings are consistent with a disentangled representation: rhythm cues are comparatively lead-robust, whereas morphology cues remain inherently lead-dependent due to both clinical criteria and acquisition variability \cite{thygesen2018udmi,gregory2021chestplacement}.

\subsection{Ranking--Decision Gap Under Protocol-Faithful Evaluation}

Across in-distribution experiments, we observe a systematic gap between ranking-oriented metrics (e.g., ROC-AUC) and fixed-threshold decision metrics (e.g., Macro F1). Under subject-aware partitioning and a fixed decision rule, ECG-RAMBA maintains strong discrimination while producing more conservative classification outcomes.

We interpret this ranking--decision gap as a consequence of deployment-oriented evaluation rather than a modeling deficiency. Subject-aware splitting reduces patient-specific leakage effects, and fixed thresholds avoid optimistic validation-time tuning that can inflate reported decision metrics. More broadly, evaluation designs that do not reflect the intended deployment setting (e.g., a new hospital or device domain) can systematically overestimate generalization \cite{leinonen2024multisourcecv}. From a clinical decision-support perspective, conservative probability assignments on ambiguous records can be preferable to overconfident hard decisions, because clinicians can select operating points aligned with clinical utility and risk tolerance (e.g., emphasizing sensitivity vs.\ specificity) \cite{vancalster2025performance}.

\subsection{Relation to Recent Zero-Shot and Multimodal ECG Models}

Recent work has pursued zero-shot robustness through large-scale pretraining and foundation-model paradigms, including multimodal ECG--text alignment and report-guided representation learning. These approaches can provide strong transfer and improved interpretability via semantic representations, often leveraging large clinical corpora and auxiliary supervision \cite{tian2024ked,mckeen2025ecgfm,tolerantecgs2025}.

ECG-RAMBA addresses a complementary (and deliberately constrained) setting: ECG-only inference under a strictly protocol-faithful regime that excludes subject overlap, threshold optimization, and test-time adaptation. While this choice yields more conservative absolute decision metrics by design, it enables a controlled examination of robustness mechanisms and failure modes (e.g., rhythm vs.\ morphology, lead sensitivity) that can be difficult to isolate in prompt-driven or multimodal pipelines. In this sense, multimodal foundation models emphasize breadth and semantic alignment, whereas ECG-RAMBA prioritizes reliability and deployment-relevant behavior under minimal assumptions \cite{mckeen2025ecgfm}.

\subsection{Synergy of Deterministic and Learnable Features}

Ablation analyses provide evidence that ECG-RAMBA benefits from combining deterministic and learnable components. Removing contextual sequence modeling consistently reduces Macro F1, indicating that stable morphology extraction alone is insufficient to integrate window-level evidence into record-level decisions. Similarly, excluding HRV features produces consistent declines, suggesting that explicit rhythm statistics remain informative even in the presence of powerful long-range backbones.

We interpret these deterministic cues not as competing predictors but as complementary constraints: MiniRocket anchors morphology in a training-stable feature space, while HRV anchors rhythm in clinically meaningful global descriptors. This combination encourages the backbone to integrate local waveform evidence with long-range temporal context, helping explain the improved robustness observed in zero-shot transfer.

\subsection{Limitations and Future Work}

ECG-RAMBA also has limitations. First, lead-set dependency for localized morphology-dominant conditions suggests that standard 12-lead training alone may not ensure spatially invariant generalization, especially under electrode misplacement or systematic lead differences across sites. Future work may explore lead-aware augmentation, spatial normalization, or carefully controlled adaptation strategies to mitigate acquisition-induced morphology shifts \cite{finlay2015electrode,gregory2021chestplacement}.

Second, fixed MiniRocket kernels provide stability and efficiency but may lack expressiveness for extremely rare or subtle morphological phenotypes. Hybrid approaches with partially learnable morphology extractors are a promising direction. Finally, although linear-time backbones support long-range modeling, deployment on wearable or edge devices may still require quantization and compression; evaluating such optimizations under strict protocol-faithful criteria remains important for real-world translation.

\section{Conclusion}
\label{sec:conclusion}

We presented ECG-RAMBA, a physiologically motivated ECG classification framework that combines deterministic morphology extraction, explicit rhythm (HRV) modeling, and efficient long-range contextual refinement. Under a protocol-faithful evaluation without subject leakage, threshold tuning, or test-time adaptation, ECG-RAMBA demonstrates improved zero-shot robustness across heterogeneous datasets. Our results highlight a consistent clinical pattern: rhythm-related evidence transfers more reliably across domains, whereas morphology-dominant conditions remain sensitive to lead configuration and acquisition differences. Together, these findings support the value of explicitly disentangling physiological cues and evaluating models under deployment-oriented protocols to better characterize robustness and failure modes in real-world ECG analysis.

\bibliographystyle{named}
\balance
\bibliography{ijcai19.bib}

\end{document}